\documentclass{article}
\usepackage{ijcai19}

\usepackage{times}
\usepackage{soul}
\usepackage{url}
\usepackage[hidelinks]{hyperref}
\usepackage[utf8]{inputenc}
\usepackage[small]{caption}
\usepackage{graphicx}
\usepackage{amsmath}
\usepackage{amsfonts}
\usepackage{booktabs}
\usepackage{algorithm}
\usepackage{algorithmic}
\usepackage{float}

\title{Disentangled Deep Autoencoding Regularization for Robust Image Classification}

\author{
Zhenyu Duan$^{1}$\and
Martin Renqiang Min$^2$\and
Li Erran Li$^{3}$\\
Mingbo Cai$^4$\and
Yi Xu$^{1}$\and
Bingbing Ni$^{1}$\\
\affiliations
$^1$Shanghai Jiao Tong University\\
$^2$NEC Labs America - Princeton\\
$^3$Pony.ai\\
$^4$Princeton University\\
}

\begin{document}

\maketitle

\begin{abstract}
  In spite of achieving revolutionary successes in machine learning, deep convolutional neural networks have been recently found to be vulnerable to adversarial attacks and difficult to generalize to novel test images with reasonably large geometric transformations. Inspired by a recent neuroscience discovery revealing that primate brain employs disentangled shape and appearance representations for object recognition, we propose a general disentangled deep autoencoding regularization framework that can be easily applied to any deep embedding based classification model for improving the robustness of deep neural networks. Our framework effectively learns disentangled appearance code and geometric code for robust image classification, which is the first disentangling based method defending against adversarial attacks and complementary to standard defense methods. Extensive experiments on several benchmark datasets show that, our proposed regularization framework leveraging disentangled embedding significantly outperforms traditional unreguarlized convolutional neural networks for image classification on robustness against adversarial attacks and generalization to novel test data.
\end{abstract}

\section{Introduction}
Deep convolutional neural networks (CNNs) have achieved revolutionary successes in many machine learning tasks such as image classification~\cite{He2016DeepRL}, object detection~\cite{nips2015_faster_rcnn}, image segmentation~\cite{He2017MaskR}, image captioning~\cite{cvpr2015_caption}, video captioning~\cite{V2T2018}, text-to-video synthesis~\cite{T2V2018}, etc. In spite of these successes, CNNs have recently been shown to be vulnerable to adversarial images which are created by adding tiny perturbations to legitimate input images~\cite{papernot2016limitations,kurakin2016adversarial}. These adversarial images that are indistinguishable from real input images by human observers can cause CNNs to output any target prediction that the attacker wants to have, suggesting that the representations learned by CNNs still lack the robustness and the capability of generalization that biological neural systems exhibit. This security issue of CNNs poses serious challenges in applying deep learning to mission-critical applications such as autonomous driving.

Various efforts have been made recently to improve the situation. Theoretical approaches such as ~\cite{icml2018_kolter} propose to learn deep ReLU-based classifiers that are provably robust against norm-bounded adversarial perturbations on the training data. However, the method still can not scale to realistic network sizes. Several empirical defense methods have been proposed~\cite{iclr2018_hotadv,iclr2018_inputtxsform,iclr2018_actprune}. Nevertheless, as shown in ~\cite{icml2018_wagner}, many of these approaches do not have guaranteed defenses. On the other hand, neuroscience inspired approaches such as ~\cite{hinton2018matrix} leverage capsule architecture to better learn the relationships between object parts and the whole, and make the model more robust and better generalizable.

Recent findings in neuroscience show that, monkey inferotemporal (IT) cortex (the final stage of the ``what" pathway of human and monkey visual system) contains separate regions where neurons exhibit disentangled representations of shape and appearance properties of face images, respectively. This disentangled representation has been applied to construct generative models of images ~\cite{deformable2018_zhu}, in which sampled disentangled embedding vectors are passed to a generator (decoder) to generate images. 

In this paper, we take the above neuroscience inspiration to regularize the embedding layer of a traditional CNN classifier by a disentangled deep autoencoder for image classification. 
In the regularizing deep autoencoder, an encoder disentangles an input image into a low-dimensional appearance code capturing the texture or background information and a low-dimensional geometric code capturing the category-sensitive shape information, and two decoders reconstruct the original high-dimensional input image from the disentangled low-dimensional code. 

Our proposed regularization framework improves CNNs' robustness and generalization mainly due to the following two reasons: First, due to the large representational capability of traditional CNNs, there are a lot of embedding functions including many pathological ones from high-dimensional input image space to the low-dimensional embedding space that can equally well minimize the classification loss; the reconstruction loss of the deep autoencoder imposes strong regularization on the low-dimensional embedding code directly used for classification; Second, the disentanglement in the embedding space imposes additional regularization on the low-dimensional code, resulting in specialized clean appearance code and geometric code, respectively, which might be, respectively, more discriminative for different classification tasks than a low-dimensional code containing a lot of information mixed with noise.

Our proposed regularization framework significantly improves the robusness of CNNs against both white-box attacks~\cite{iclr2015_fgsm,iclr_worshop2017_bim,papernot2016limitations,papernot2016distillation,carlini2017towards}, which assume the model is completely known, and black-box attacks~\cite{cvprw2017_blackbox,asia_ccs2017_Papernot}, which  can only query the model for a limited number of times.

Our contributions in this paper are as follows: (i) We propose the first disentangling based deep autoencoding regularization framework to improve the robustness of CNNs, which can be easily applied to any embedding based CNN classification model and combined with standard defense methods; (ii) Extensive experiments show that, our regularization framework dramatically improves the robustness of standard CNNs against the attacks from adversarial examples; (iii) Our proposed framework helps CNNs learn more robust image representation and gain better generalization to novel test data with large geometric differences, achieving state-of-the-art generalization ability compared with capsule networks \cite{hinton2018matrix}.

\section{Method}
\subsection{Disentangled Deep Autoencoding Regularization}
\label{sec:model}
\begin{figure}[ht]
\centering
\includegraphics[scale=0.35]{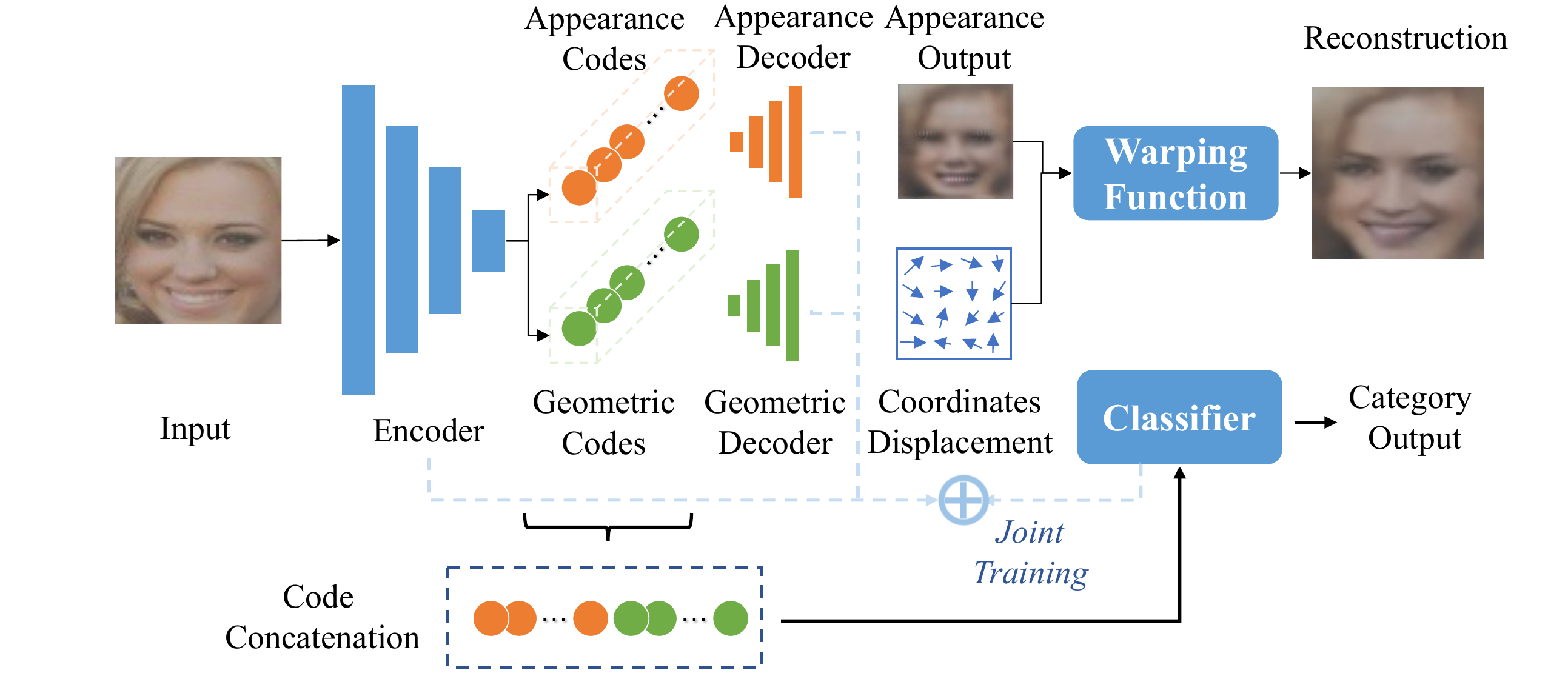}
\caption{The disentangled deep autoencoding regularization framework for classification.}
\label{fig:arch}
\end{figure}
Suppose we have a $K$-class training set $D_T = \{(\mathbf{I}^i, \mathbf{y}^i), i=1,\cdots,N\}$ containing N pairs of samples, where $\mathbf{I}^i \in \mathbb{R}^{H \times W \times 3}$ is the image to be classified and $\mathbf{y^i} \in \{0,\cdots,K-1\}$ is the corresponding label. Given a new test image $\mathbf{I^j}$ that does not belongs to $D_T$, the classification task is to predict the correct label $\mathbf{y^j}$ based on the features $\mathbf{C^j}$ extracted from $\mathbf{I^j}$.

To encourage learning visual features robust to natural variations of images, we employ additional learning objective and augmentation task to force a classifier to learn disentangled geometric code and appearance code.
As shown in Figure~\ref{fig:arch}, the proposed framework contains two parts. The first part is a regularizing autoencoder with disentangled feature embedding, displayed on the top. The second part performs classification on the concatenated embedding feature vector, displayed on the bottom.
For any image $\mathbf{I^i}$, the encoder maps it to a pair of appearance and geometric codes as follows,
\begin{equation}
    \mathbf{C}_a, \mathbf{C}_g = \textbf{\text{Encoder}}(\mathbf{I}),
\end{equation}
in which our encoder learns a disentangled embedding of the input image, namely the appearance code $\mathbf{C}_a$ and the geometric code $\mathbf{C}_g$. The appearance decoder or generator takes the appearance code to generate the appearance output $\mathbf{X}_a$. The geometric decoder takes the geometric code and generates coordinate displacement of the appearance output $\mathbf{X}_g$.
\begin{equation}
    \mathbf{X}_a = \textbf{\text{Decoder\_A}}(\mathbf{C}_a),
    \mathbf{X}_g = \textbf{\text{Decoder\_G}}(\mathbf{C}_g).
\end{equation}
The warping function $\mathbf{F}_{W}$ warps the appearance with the coordinate displacement generated by the geometric decoder and generates the reconstruction output $\mathbf{X}$. This will be described in detail later.
\begin{equation}
    \label{Reconstruction}
    \mathbf{X} = \mathbf{F}_W(\mathbf{X}_a,\mathbf{X}_g).
\end{equation}
The classifier $\mathbf{F}_{CLS}$ takes both the appearance code and geometric code as input and predicts category probability $\hat{\mathbf{y}}$.
\begin{equation}
    \hat{\mathbf{y}} = \mathbf{F}_{CLS}(\mathbf{C}_a, \mathbf{C}_g).
\end{equation}

In training process, the encoder, two decoders as well as classifier are trained jointly and the loss function $\mathcal{L}$ is defined as
\begin{equation}
    \mathcal{L} = \mathcal{L}_{crossentropy}(\hat{\mathbf{y}}, \mathbf{y}) + \lambda \Vert \mathbf{X} - \mathbf{I}\Vert,
\end{equation}
where $\mathbf{y}$ is the training label vector, $\Vert \mathbf{X} - \mathbf{I}\Vert$ is the reconstruction error (frobenius norm), and $\mathcal{L}_{crossentropy}$ is the cross-entropy loss, and $\lambda$ is the regularization weight for the reconstruction loss.

The disentangled representation is learned in an unsupervised manner. The warping function naturally encourages the separation of appearance code from geometric code because $\mathbf{X}_g$ represents coordinate displacement of $\mathbf{X}_a$.

Note that the complete architecture is needed only for training. For inference, the two decoders, warping function and the reconstruction are not needed.

\subsubsection{Warping Function}

The warping function, such as the one used in Spatial Transformer networks (STN) \cite{jaderberg2015spatial}, is usually a combination of geometric transformation for pixels and interpolating operation.
The geometric transformation usually consists of two  types: affine transformation and pixel displacement transformation (like the optical flow field). We employ the second type in this work, similarly as in ~\cite{deformable2018_zhu}. 
The warping function takes the displacement field and the "unwarped" image as the input. The vector at each pixel of the displacement field is defined to be related with the spatial shift needed to move the destination pixel ($x$,$y$) to source pixel ($x_0$,$y_0$) in the output of the appearance generator as below:
\begin{equation}
    \mathbf{\Phi_g}
    \left(\begin{matrix}
    x\\y
    \end{matrix}\right)=
    \left(\begin{matrix}
    x+dx\\y+dy
    \end{matrix}\right)=
    \left(\begin{matrix}
    x_0\\y_0
    \end{matrix}\right).
\end{equation}
where $\mathbf{\Phi_g}$ denotes the pixel displacement operation.

The outputs of the pixel displacement operation are usually decimals. In order to get a set of pixels whose coordinates are integers, we employ the differentiable bi-linear interpolation in this work. The bi-linear interpolation process can be represented as:
\begin{eqnarray}
    \mathbf{X}(x,y) = & \Sigma_{h=1}^H\Sigma_{w=1}^W \mathbf{X}_a(h,w) M(1-\vert x_0 -w\vert)\nonumber \\
      & M(1-\vert y_0-h\vert).
\end{eqnarray}
where $M(\cdot)$ means the function $\text{max}(0,\cdot)$, and $H$, $W$ are the height and width of the input image respectively.

In this way, various pictures of the same object can be encoded with similar appearance code, and their differences in viewing angles, locations, and deformation are captured by the geometric code. With this idea, we use data augmentation detailed below to enforce the separation of appearance and geometric information.  

\subsection{Data Augmentation and Code Regularization}
\label{subsec:aug}
In this section, we will describe how to combine the proposed regularization framework with deformation-based data augmentation method.
This deformation-based data augmentation is achieved by random 2D-stretching, scaling and spatial shift of the target images. To encourage appearance code to capture information invariant to spatial transformation, we impose L2 regularization on the distance between the appearance codes of images before (denoted as $\mathbf{I}$) and after deformation ($\mathbf{I'}$).

The appearance code and geometric code of $\mathbf{I'}$ from encoder are $\mathbf{C}'_a$ and $\mathbf{C}'_g$. And the corresponding reconstruction output and classification output are denoted as $\mathbf{X}'$ and $\hat{\mathbf{y}}'$. The new loss function $\mathcal{L}_{AUG}$ is 
\begin{equation}
\begin{split}
    \mathcal{L}_{AUG} = \mathcal{L} &+  \mathcal{L}_{crossentropy}(\hat{\mathbf{y}}', \mathbf{y})+ \lambda \Vert \mathbf{X}' - \mathbf{I}'\Vert\\
    &+\gamma \Vert \mathbf{C}_a - \mathbf{C}'_a\Vert.
\end{split}
\end{equation}
where $\gamma$ is the scale weight for appearance code regularization. We set $\lambda$ and $\gamma$ as 1 in experiments.

As shown in Section~\ref{sec:exp}, data augmentation and code regularization further encourages the separation of appearance code from geometric code. This leads to much improved generalization to novel test data with large geometric transformation.

\section{Experiments}
\label{sec:exp}
In this section, the robustness and generalization to novel test data of the proposed regularization framework is evaluated. In addition, ablation studies are performed to analyze the individual mechanisms that contribute to the gains. 

\subsection{Robustness Against Adversary Attack}
We evaluate our framework's robustness against both white-box and black-box attacks. White-box attacks assume that the adversary knows the network architecture and the model parameters after training or has access to labeled training data. An adversary can use the detailed information to construct a perturbation for a given image. Black-box attacks only rely on a limited number of queries to the model to construct an adversarial image.

To compare with the related work such as ~\cite{hinton2018matrix}, we use three dataseets, the MNIST dataset, the FaceScrub dataset, and the CIFAR10 datasset. The MNIST dataset of handwriten digits has 60000 for training and 10000 for test. The FaceScrub data set contains 106,863 face images of 530 celebrities taken under real-world conditions. The CIFAR10 dataset contains 60,000 color images with a resolution of 32x32 in 10 different classes, in which 50,000 are used for training and 10,000 are used for test.

\subsubsection{White-Box Attack}
We evaluate our model's robustness using popular white box attack methods: fast gradient sign method (FGSM)~\cite{iclr2015_fgsm} and Basic Iterative Method (BIM)~\cite{iclr_worshop2017_bim}.  FGSM computes the gradient of the loss with respect to each pixel intensity and then changes the pixel intensity by a fixed amount $\epsilon$ in the direction that increases the loss.  BIM applies FGSM with multiple smaller steps when creating the adversarial image. FGSM and BIM are adopted as the main attack methods by a series of works. Since these methods attack the image based on the gradient, they pose a great threat to modern CNN architecture trained through gradient back-propagation.

We use the same benchmark CNN architecture as ~\cite{hinton2018matrix}. It has 4 convolutional layers with a softmax output. For fair comparison, the classifier part of the proposed framework uses exactly the same architecture as the benchmark CNN architecture. To test the robustness of the proposed framework, adversarial images are generated from the test set using a fully trained model. We then reported the accuracy of the model on these images. All models are trained without adversary examples. 

As show in Table~\ref{mnistFGSM} and ~\ref{mnistBIM}, the proposed regularization framework is much more robust than the benchmark CNN architecture. \footnote{The visualization of feature embedding of legitimate images and adversarial images on the MNIST dataset can be found in the supplementary material.}
\begin{table}[ht]
\centering
\caption{Attack Success Rate Comparison on MNIST Dataset (FGSM)}
\label{mnistFGSM}
\begin{tabular}{cccccc}
\hline
$\epsilon$ & 0.1& 0.2& 0.3& 0.4& 0.5\\ \hline
Base CNN & 76.3\% &89.4\% & 96.7\% & 97.3\% & 97.6\% \\
Ours & \textbf{8.8\%} & \textbf{12.3\%} & \textbf{16.2\%} & \textbf{25.8\%} & \textbf{38.5\%} \\ \hline
\end{tabular}
\end{table}

\begin{table}[ht]
\centering
\caption{Attack Success Rate Comparison on MNIST Dataset (BIM)}
\label{mnistBIM}
\begin{tabular}{cccccc}
\hline
$\epsilon$ (iters) & 0.01   & 0.02   & 0.03   & 0.04   & 0.05   \\ \hline
Base CNN (2)   & \textbf{7.4\%} & 29.4\% & 54.3\% & 74.4\% & 87.6\% \\
Ours (2)        & 9.8\% & \textbf{20.6\%} & \textbf{27.4\%} & \textbf{31.0\%} & \textbf{33.4\%} \\ \hline
Base CNN (3)   & \textbf{7.8\%} & 30.7\% & 55.4\% & 75.3\%  & 88.3\%  \\
Ours (3)        & 10.3\% & \textbf{22.1\%} & \textbf{29.2\%} & \textbf{32.4\%} & \textbf{34.7\%} \\ \hline
\end{tabular}
\end{table}

It's worth noting that for defending against the adversary attack with large perturbation, simply adding the proposed regularization even exceeds some defense models that used adversary training. For example, our model shows a higher performance (61.5\%) over the gradient-regularized model \cite{ross2017improving} (lower than 60\%) when perturbed with a big perturbation (0.5) by FGSM on MNIST Dataset.

We further evaluate our model's robustness using FaceScrub dataset. To simplify the problem, we randomly select 50 faces for the classification task. For the benchmark CNN architecture, the training and test accuracy is 99.9\% and 74.2\%, respectively. While after adding the proposed regularization, the training and test accuracy is 99.9\% and 79.8\%, respectively. As shown in Table~\ref{faceFGSM} and ~\ref{faceBIM}, the regularization framework is much more robust than the benchmark CNN model.
Please note that $\epsilon = 0.05$ is big enough as a perturbation in this task.
\begin{table}[ht]
\centering
\caption{Attack Success Rate Comparison on FaceScrub dataset (FGSM)}
\label{faceFGSM}
\begin{tabular}{cccccc}
\hline
$\epsilon$   & 0.01   & 0.02   & 0.03   & 0.04   & 0.05   \\ \hline
Base CNN  & 57.8\% & 67.0\% & 72.9\% & 78.3\% & 82.2\% \\
Ours      & \textbf{19.9\%} & \textbf{24.6\%} & \textbf{27.9\%} & \textbf{33.4\%} & \textbf{38.9\%} \\ \hline
\end{tabular}
\end{table}

\begin{table}[ht]
\centering
\caption{Attack Success Rate Comparison on FaceScrub dataset (BIM)}
\label{faceBIM}
\begin{tabular}{cccccc}
\hline
$\epsilon$   & 0.01   & 0.02   & 0.03   & 0.04   & 0.05   \\ \hline
Base CNN  & 72.1\% & 94.7\% & 99.2\% & 99.8\% & 99.6\% \\
Ours      & \textbf{26.2\%} & \textbf{35.6\% }& \textbf{44.1\%} & \textbf{52.7\%} & \textbf{60.7\%} \\ \hline
\end{tabular}
\end{table}

Finally, the framework is tested with CIFAR10 dataset to evaluate the proposed disentangling method on general natural images where the information of categories can lie in both appearance code and geometric code in varying degrees. In this experiment, data augmentation methods including random cropping and random flipping are applied to the normalized input data. We use 110-layer ResNet \cite{he2016deep} as the baseline and it achieves 8.5\% error rate on the test set. To add the proposed regularization, the output of the final fully-connected layer of the ResNet is modified to generate geometric and appearance code. The decoder is the same as described in the main text. With our proposed special regularization, the network achieves 8.2\% error rate. The defense result is shown as below:
\begin{table}[ht]
\centering
\caption{Attack success rate comparison on CIFAR10 dataset (FGSM)}
\label{cifaFGSM}
\begin{tabular}{cccccc}
\hline
 $\epsilon$           & 0.1    & 0.2    & 0.3    & 0.4    & 0.5    \\ \hline
Base CNN    & 29.5\% & 39.9\% & 47.4\% & 53.8\% & 59.3\% \\
Ours        & \textbf{19.0\%} & \textbf{26.8\%} & \textbf{33.5\%} & \textbf{39.4\%} & \textbf{44.4\%} \\ \hline
\end{tabular}
\end{table}

\begin{table}[ht]
\centering
\caption{Attack success rate comparison on CIFAR10 dataset (BIM)}
\label{cifaBIM}
\begin{tabular}{cccccc}
\hline
$\epsilon$ (iters)       & 0.1    & 0.2    & 0.3    & 0.4    & 0.5    \\ \hline
CNN Baseline (2) & 0.403          & 0.615         & 0.761          & 0.844          & 0.884          \\
Ours (2)         & \textbf{0.247} & \textbf{0.502} & \textbf{0.700} & \textbf{0.808} & \textbf{0.867} \\ \hline
CNN Baseline (3) & 0.474          & 0.706          & 0.833          & 0.891          & 0.908          \\
Ours (3)         & \textbf{0.346} & \textbf{0.626} & \textbf{0.783} & \textbf{0.851} & \textbf{0.892} \\ \hline
\end{tabular}
\end{table}

As shown in Table~\ref{cifaFGSM} and Table~\ref{cifaBIM}, our method outperforms the CNN baseline for the FGSM attack and BIM attack. The proposed method improves over CNN baseline by 15\% to 37\% for FGSM attack.

\subsubsection{Black-Box Attack}
The first black-box attack method we use is proposed in ~\cite{cvprw2017_blackbox}. The basic idea is greedy local search. It is an iterative search procedure, where in each round a local neighborhood is used to refine the current image and to optimize some objective function depending on the network output. The number of iterations of the black-box attack is adjusted as an experimental variable. As shown in Table~\ref{blackbox:greedy}, the proposed regularization method can significantly promote the benchmark CNN architecture.
\begin{table}[ht]
\centering
\caption{Attack Success Rate on MNIST dataset (Local Search)}
\label{blackbox:greedy}
\begin{tabular}{ccc}
\hline
Iters     & 50     & 150    \\ \hline
Base CNN  & 57.3\% & 62.5\%   \\
Ours      & \textbf{47.9\%} & \textbf{57.8\%}\\
\hline
\end{tabular}
\end{table}

Besides the greedy local search black-box attack approach, another black-box attack approach~\cite{asia_ccs2017_Papernot} leverages \textit{transferability} property in creating adversarial examples. The attack first creates attack examples using a known deep learning model (white-box attack). The attack then apply these adversarial examples on the deep learning model with no access to model architecture and parameters. We tried this type of black-box attack and observed that our method has no noticeable advantage over benchmark CNN architecture. This is not surprising as transferability property applies to both deep learning models and non-deep learning models such as decision trees and SVM in general~\cite{asia_ccs2017_Papernot}.

\subsubsection{Joint Defense}
\label{advTrain}
As has been stated, the proposed framework is orthogonal to other defense techniques. It can potentially be combined with these defense techniques to further improve the robustness of CNN models. As an example, we combine it with adversarial training. Specifically, we train the model with legitimate inputs (MNIST) and adversarial samples generated by the model using FGSM attack method with $\epsilon$ of 0.3~\cite{ross2017improving}. 

\begin{table}[ht]
\centering
\caption{Attack Success Rate Comparison on MNIST Dataset (FGSM with adversarial training)}
\label{advT}
\begin{tabular}{cccccc}
\hline
$\epsilon$     & 0.1    & 0.2    & 0.3    & 0.4    & 0.5    \\ \hline
Base CNN  & 9.55\% & 22.0\% & 32.9\% & 42.9\% & 55.4\% \\
Ours         & \textbf{9.49\%} & \textbf{12.4\%} & \textbf{15.6\%} & \textbf{22.3\%} & \textbf{34.8\%} \\ \hline
\end{tabular}
\end{table}
As shown in Table~\ref{advT}, the result proves that our regularization method could be combined with adversarial training to further improve the robustness of the classifiers. Our method with no adversarial training (see Table 1 in the main paper) outperforms CNN baseline with adversarial training (See Table~\ref{advT}). Our method with adversarial training (See Table~\ref{advT}) further outperforms our method without adversarial training (see Table 1 in the main paper). For example, it is 34.8\% vs 38.5\% for $\epsilon$ = 0.5.

\subsection{Generalization}
\label{generalizaiton}
We evaluate our model's generalization to novel test data with large geometric transformation on the smallNORB dataset\cite{lecun2004learning}. The dataset contains five categories of 3D objects (animals, human, airplanes, trucks and cars), each with 10 instances. These images are captured through a combination of six illumination conditions, nine elevations, and eighteen azimuths and have a resolution of 96 $\times$ 96. The training set and the testing set contain 5 different instances of each category. This data set requires the model to learn the common features within each target category.


For ease of comparison, we use the same experimental setting and the same network architecture of the benchmark CNN as reported in \cite{hinton2018matrix}. Both our model and baseline CNN model are trained on one-third of the training set, which contains six smallest azimuths or 3 smallest elevations, and tested on two-thirds of the test set, which contains images of totally different azimuths and elevations from those in training set. For training, we randomly crop $48 \times 48$ sub-patch from the input image, resize the patch into $32 \times 32$ and add small random jitters to brightness and contrast. For testing, the samples are directly cropped from the center. For both training and testing, all data have been normalized to have zero mean and unit variance.

To better disentangle geometric code from appearance code, we use the data augmentation and code regularization technique proposed in Section~\ref{subsec:aug}. Specifically, deformations such as displacement and stretching to the inputs is applied. The extent of stretching in this experiment is randomly selected from \{2, 4, 6, 8\} pixels, the distance of displacement in each dimension (in the range of \{2, 4, 6, 8\}) and the orientation of displacement are also randomly selected. Then the samples are re-sized back to 32 $\times$ 32 and padded by pixels at the edge when necessary. These deformation operations do not change the appearance contents. The L2 norm of the difference between appearance codes of an input image and those corresponding to its deformed image are used as a regularization term as in Section~\ref{subsec:aug}.  
\begin{table}[h]
\centering
\caption{Generalization comparison on novel viewpoints}
\label{genNORB}
\begin{tabular}{ccc}
\hline
Test set  & Azimuth & Elevation \\ \hline
Base CNN  & 84.8\%  & 78.8\%    \\
Ours      & \textbf{86.6\%}  & \textbf{85.3\%}    \\ \hline
\end{tabular}
\end{table}

\begin{figure}[ht]
\centering
\includegraphics[width=\linewidth]{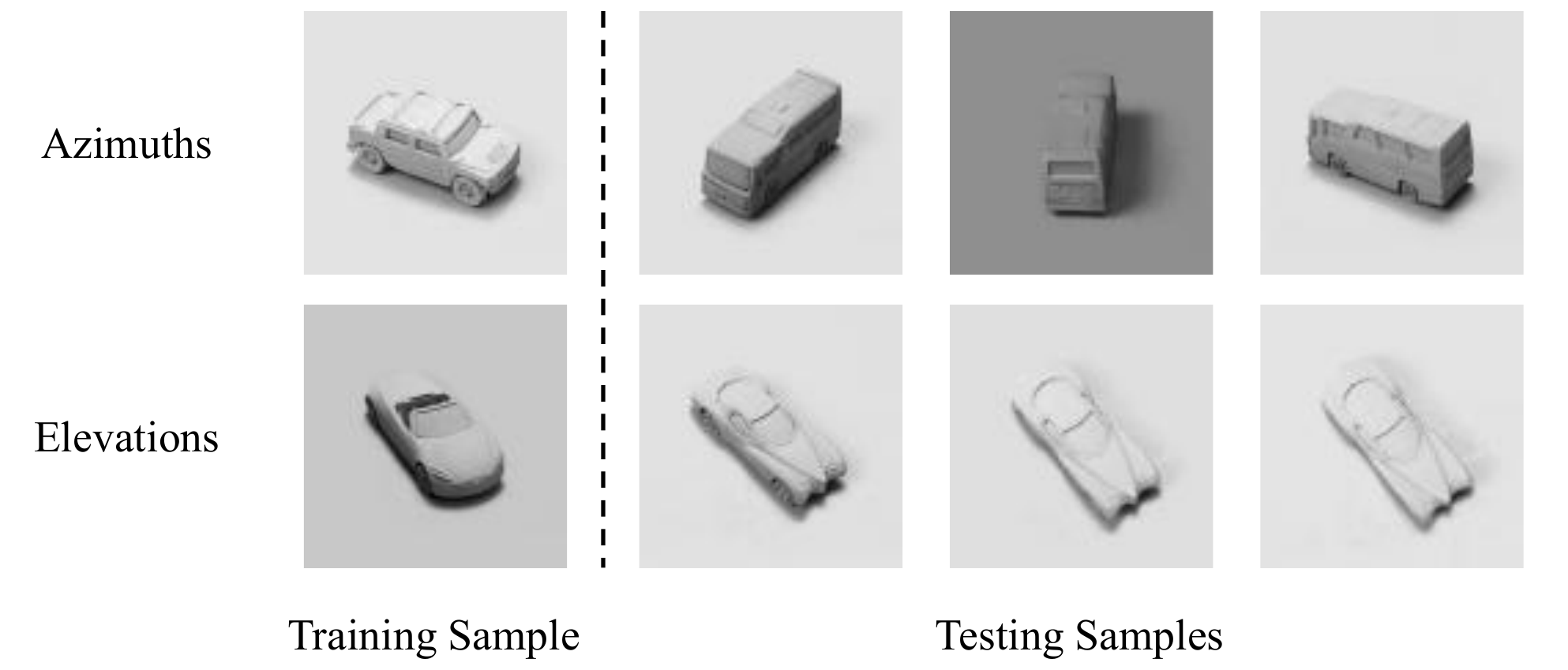}
\caption{The comparison of distribution between the training data and testing data of the smallNORB dataset's. The top row shows trucks under different azimuths in training and testing set, while the second row displays the cars under different elevations. Notice that the instances in training set and testing set are highly different.}
\label{fig:norbSample}
\end{figure}

As shown in Table~\ref{genNORB}, our model is 8\% better in accuracy when generalizing to novel elevation and 2\% better in case of novel azimuth. We note that of particular challenge, the task is to generalize to novel elevation and azimuth. The test distribution is different from the training distribution which makes the classification extremely hard. 
For example, Figure~\ref{fig:norbSample} shows the cars in testing set with the highest visual similarity to example cars in training set. With largely different azimuths and elevations and even shapes, the generalization task can be difficult even for human observers.

\begin{figure*}[h]
\centering
\includegraphics[width=0.7\linewidth]{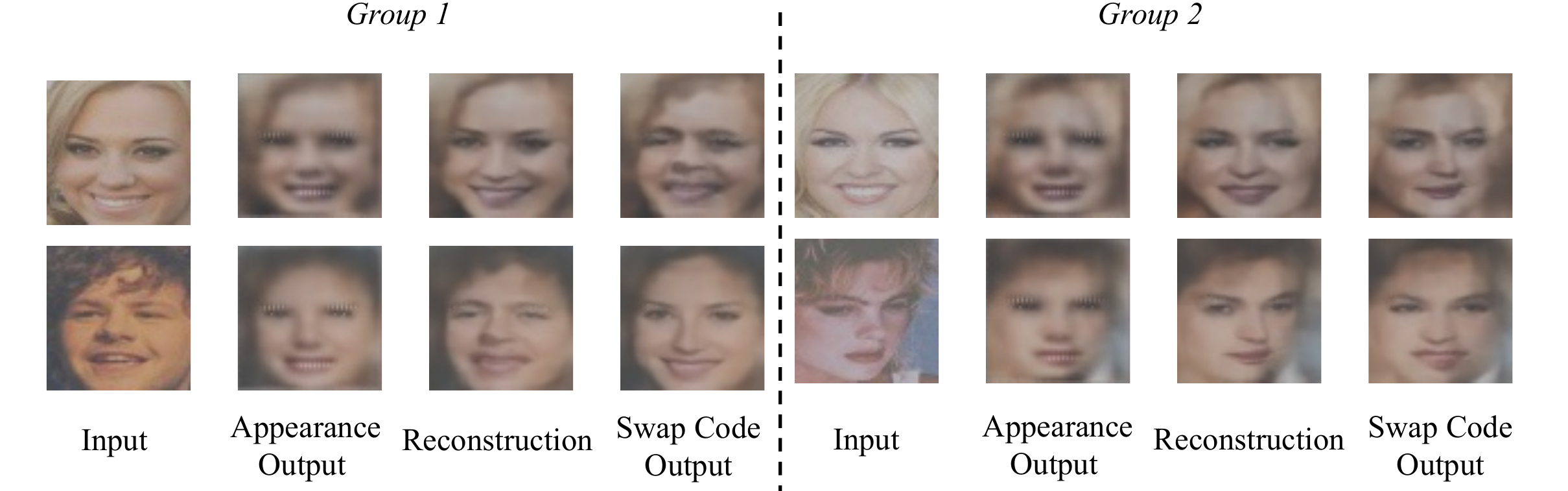}
\caption{The synthesis study on code representation. Appearance outputs $\mathbf{Decoder\_A}(\mathbf{C}_a)$ are all front-facing faces. Exchanging appearance and geometric codes generates new images with exchanged properties of the original images, as shown in columns of \textit{swap code output}. Group 1 displays swap-code outputs with appearance code exchanged; Group 2 displays swap-code outputs with geometric code exchanged.}
\label{fig:swap}
\end{figure*}
Note that our CNN baseline performance on novel elevation is lower than that reported in ~\cite{hinton2018matrix} while the performance on novel azimuth is higher. Although we use exactly the same network architecture according to the description in ~\cite{hinton2018matrix}, such difference may be unavoidable given that many parameters such as brightness and contrast jittering range need to be tuned, which were not mentioned in the paper. 
Our result on generalizing to novel azimuth achieves state-of-the-art. We achieve 86.6\%, comparable to 86.5\%  of CapsuleNet as reported in~\cite{hinton2018matrix}. For generalizing to novel elevation, even compared with the CNN benchmark result of 82.2\% reported in ~\cite{hinton2018matrix}, our result of 85.3\% is still 3.7\% better. It is worth noting that our method is orthogonal to the capsule technique ~\cite{hinton2018matrix}. Potentially the strength of feature disentangling and the capsule design may be combined to further improve performance.

\subsection{Synthesis Study}
To evaluate the quality of disentangling, we examine the property of "unwarped" face image $\mathbf{X}_a$ and the exchangeability of $\mathbf{C}_g$ and $\mathbf{C}_a$.
Two faces ($\mathbf{I}^0$ and $\mathbf{I}^1$) are randomly selected and encoded by our model as $\mathbf{C}_a^0, \mathbf{C}_g^0)$ and $(\mathbf{C}_a^1, \mathbf{C}_g^1)$. We then exchange their codes into $(\mathbf{C}_a^0, \mathbf{C}_g^1)$ and $(\mathbf{C}_a^1, \mathbf{C}_g^0)$. The swapped codes are then decoded by the appearance and geometric decoder and new images are synthesized by the warping function: $\mathbf{X}^i_{new} = \mathbf{F}_W(\mathbf{Decoder\_A}(\mathbf{C}_a^i), \mathbf{Decoder\_G}(\mathbf{C}_g^{1-i}))$, where $i \in \{0, 1\}$.



Two pairs of ($\mathbf{X}^0_{new}$,  $\mathbf{X}^1_{new}$) are visualized on Figure~\ref{fig:swap}. The appearance output is $\mathbf{Decoder\_A}(\mathbf{C}_a^i)$, which is the "unwarped" image. The reconstruction output is the warped output without swapping code ($\mathbf{X}^i$) and the swap code output is the the warped output after swapping code ($\mathbf{X}^i_{new}$). Group 1 and Group 2 emphasize the effects of maintaining geometric code and appearance code, respectively, while exchanging the other code.
This figure prominently reflects the role of the geometric code and the appearance code in the face reconstruction process of the proposed model. It can be observed that the geometric code mainly contains some deformation information, such as the facial features, the expression and even the face orientation. On the other hand, the appearance code contains texture information, such as skin color, hair color, background as well as the basic structure of face.

\subsection{Ablation Studies}
\label{ablation}
We conduct ablation studies to analyze the relative contributions of each design component in our framework: disentangled representation learning, augmented training, L2 regularization of appearance code difference when an input image is deformed. 
We first investigate the benefit of disentangled feature learning in terms of robustness to FSGM attack using MNIST dataset. To disable disentangled feature learning, the warping function component is removed from Figure~\ref{fig:arch}. We refer it as 'Reconstruction' in Table ~\ref{ablation:mnist}.
As shown in Table~\ref{ablation:mnist}, disentangled feature learning is essential for the model to be robust against FGSM, e.g. 16.15\% success rate vs 43.83\% success rate at $\epsilon$=0.3.

\begin{table}[h]
\centering
\caption{Ablation Study: Attack Success Rate on the MNIST dataset (FGSM attack)}
\label{ablation:mnist}
\begin{tabular}{cccccc}
\hline
$\epsilon$     & 0.1    & 0.2    & 0.3    & 0.4    & 0.5    \\ \hline
Recon. & 11.5\% & 24.1\% & 43.8\% & 57.8\% & 67.6\% \\
Ours           & \textbf{8.8\%} & \textbf{12.3\%} & \textbf{16.2\%} & \textbf{25.8\%} & \textbf{38.5\%} \\ \hline
\end{tabular}
\end{table}

\begin{table}[h]
\centering
\caption{Ablation Study on the smallNORB dataset; AT: augmented training, CR: code regularization}
\label{ablation:norb}
\begin{tabular}{ccc}
\hline
Settings           & Base CNN & Ours            \\ \hline
Base model   & 81.3\%    & \textbf{84.4\%} \\
with AT & 84.8\%    & \textbf{85.3\%} \\
with AT and CR     & None      & \textbf{86.6\%} \\ \hline
\end{tabular}
\end{table}

Another ablation study are further conducted on generalization using smallNORB dataset.  The experimental settings is the same as section \ref{generalizaiton}. The base model in Table \ref{ablation:norb} is the architecture in Figure~\ref{fig:arch} with no augmented training and additional regularization. The augmented training settings in Table ~\ref{ablation:norb} refers to training with augmented data by deforming input data. The augmented training and code regularization setting in Table~\ref{ablation:norb} means that the L2 regularization of appearance code difference between an input image and its deformed one is further imposed.

Results in Table \ref{ablation:norb} show that with the proposed regularization framework, the CNN base model is promoted in all settings. It also shows that augmented training and additional regularization helps. With both augmented training and L2 regularization, the improvement over the base model is 2.2\%.

\section{Related Work}
There has been several works on learning disentangled feature representation in the machine learning literature.  \cite{Denton2017UnsupervisedLO} and \cite{Hsieh2018LearningTD} disentangled video representation into content and pose. 
\cite{ICLR2017_honglak} disentangled video representation into content and motion using image differences for video prediction. 
\cite{Li2018UnsupervisedDR} learn disentangled representations using analogy reasoning. Unlike previous methods that model pose with a rigid affine mapping, the coordinate displacement in our work is a non-rigid deformable mapping. In addition, most of previous methods require learning from video data in order to utilize the time-dependent features in pose and time-independent ones in content. Our model does not require time-series data and is suitable for independently sampled image data. Finally, our proposed framework aims to improve the robustness of CNNs, which has a completely different purpose from previous models.
\section{Conclusion and Future Work}
Robustness and generalization are paramount to deploying deep learning models in mission-critical applications. 
In this paper, we take inspiration from neuroscience and show that disentangling feature representation into appearance and geometric code can significantly improve the robustness and generalization of CNNs.

There are other mechanisms at play for the robustness and generalization ability of primate brain's perception. In the future, we plan to explore other neuroscience motivated directions. For example, our model can be potentially combined with capsule networks~\cite{hinton2018matrix}.

\bibliographystyle{named}
\small{
\bibliography{ijcai19}

\begin{thebibliography}{}

\bibitem[\protect\citeauthoryear{Athalye \bgroup \em et al.\egroup
  }{2018}]{icml2018_wagner}
Anish Athalye, Nicholas Carlini, and David~A. Wagner.
\newblock Obfuscated gradients give a false sense of security: Circumventing
  defenses to adversarial examples.
\newblock In {\em {ICML}}, 2018.

\bibitem[\protect\citeauthoryear{Buckman \bgroup \em et al.\egroup
  }{2018}]{iclr2018_hotadv}
Jacob Buckman, Aurko Roy, Colin Raffel, and Ian Goodfellow.
\newblock Thermometer encoding: One hot way to resist adversarial examples.
\newblock In {\em ICLR}, 2018.

\bibitem[\protect\citeauthoryear{Carlini and Wagner}{2017}]{carlini2017towards}
Nicholas Carlini and David Wagner.
\newblock Towards evaluating the robustness of neural networks.
\newblock In {\em 2017 IEEE Symposium on Security and Privacy (SP)}, pages
  39--57. IEEE, 2017.

\bibitem[\protect\citeauthoryear{Denton and
  Birodkar}{2017}]{Denton2017UnsupervisedLO}
Emily~L. Denton and Vighnesh Birodkar.
\newblock Unsupervised learning of disentangled representations from video.
\newblock In {\em NIPS}, 2017.

\bibitem[\protect\citeauthoryear{Dhillon \bgroup \em et al.\egroup
  }{2018}]{iclr2018_actprune}
Guneet~S. Dhillon, Kamyar Azizzadenesheli, Zachary~C. Lipton, Jeremy Bernstein,
  Jean Kossaifi, Aran Khanna, and Anima Anandkumar.
\newblock Stochastic activation pruning for robust adversarial defense.
\newblock 2018.

\bibitem[\protect\citeauthoryear{Goodfellow \bgroup \em et al.\egroup
  }{2015}]{iclr2015_fgsm}
Ian Goodfellow, Jonathon Shlens, and Christian Szegedy.
\newblock Explaining and harnessing adversarial examples.
\newblock In {\em ICLR}, 2015.

\bibitem[\protect\citeauthoryear{Guo \bgroup \em et al.\egroup
  }{2018}]{iclr2018_inputtxsform}
Chuan Guo, Mayank Rana, Moustapha Ciss{\'{e}}, and Laurens van~der Maaten.
\newblock Countering adversarial images using input transformations.
\newblock In {\em ICLR}, 2018.

\bibitem[\protect\citeauthoryear{He \bgroup \em et al.\egroup
  }{2016a}]{He2016DeepRL}
Kaiming He, Xiangyu Zhang, Shaoqing Ren, and Jian Sun.
\newblock Deep residual learning for image recognition.
\newblock {\em CVPR}, pages 770--778, 2016.

\bibitem[\protect\citeauthoryear{He \bgroup \em et al.\egroup
  }{2016b}]{he2016deep}
Kaiming He, Xiangyu Zhang, Shaoqing Ren, and Jian Sun.
\newblock Deep residual learning for image recognition.
\newblock In {\em Proceedings of the IEEE conference on computer vision and
  pattern recognition}, pages 770--778, 2016.

\bibitem[\protect\citeauthoryear{He \bgroup \em et al.\egroup
  }{2017}]{He2017MaskR}
Kaiming He, Georgia Gkioxari, Piotr Doll{\'a}r, and Ross~B. Girshick.
\newblock Mask r-cnn.
\newblock {\em ICCV}, pages 2980--2988, 2017.

\bibitem[\protect\citeauthoryear{Hinton \bgroup \em et al.\egroup
  }{2018}]{hinton2018matrix}
Geoffrey~E Hinton, Sara Sabour, and Nicholas Frosst.
\newblock Matrix capsules with em routing.
\newblock In {\em ICLR}, 2018.

\bibitem[\protect\citeauthoryear{Hsieh \bgroup \em et al.\egroup
  }{2018}]{Hsieh2018LearningTD}
Jun-Ting Hsieh, Bingbin Liu, De-An Huang, Li~Fei-Fei, and Juan~Carlos Niebles.
\newblock Learning to decompose and disentangle representations for video
  prediction.
\newblock In {\em NIPS}, 2018.

\bibitem[\protect\citeauthoryear{Jaderberg \bgroup \em et al.\egroup
  }{2015}]{jaderberg2015spatial}
Max Jaderberg, Karen Simonyan, Andrew Zisserman, et~al.
\newblock Spatial transformer networks.
\newblock In {\em NIPS}, pages 2017--2025, 2015.

\bibitem[\protect\citeauthoryear{Kurakin \bgroup \em et al.\egroup
  }{2016}]{kurakin2016adversarial}
Alexey Kurakin, Ian Goodfellow, and Samy Bengio.
\newblock Adversarial examples in the physical world.
\newblock {\em arXiv preprint arXiv:1607.02533}, 2016.

\bibitem[\protect\citeauthoryear{Kurakin \bgroup \em et al.\egroup
  }{2017}]{iclr_worshop2017_bim}
Alexey Kurakin, Ian Goodfellow, and Samy Bengio.
\newblock Adversarial examples in the physical world.
\newblock {\em ICLR Workshop}, 2017.

\bibitem[\protect\citeauthoryear{LeCun \bgroup \em et al.\egroup
  }{2004}]{lecun2004learning}
Yann LeCun, Fu~Jie Huang, and Leon Bottou.
\newblock Learning methods for generic object recognition with invariance to
  pose and lighting.
\newblock In {\em CVPR}, 2004.

\bibitem[\protect\citeauthoryear{Li \bgroup \em et al.\egroup
  }{2018a}]{T2V2018}
Yitong Li, Martin~Renqiang Min, Dinghan Shen, David~E. Carlson, and Lawrence
  Carin.
\newblock Video generation from text.
\newblock In {\em Proceedings of the Thirty-Second {AAAI} Conference on
  Artificial Intelligence, New Orleans, Louisiana, USA}, pages 7065--7072,
  2018.

\bibitem[\protect\citeauthoryear{Li \bgroup \em et al.\egroup
  }{2018b}]{Li2018UnsupervisedDR}
Zejian Li, Yongchuan Tang, and Yongxing He.
\newblock Unsupervised disentangled representation learning with analogical
  relations.
\newblock In {\em IJCAI}, 2018.

\bibitem[\protect\citeauthoryear{Maaten and
  Hinton}{2008}]{maaten2008visualizing}
Laurens van~der Maaten and Geoffrey Hinton.
\newblock Visualizing data using t-sne.
\newblock {\em Journal of machine learning research}, 9(Nov):2579--2605, 2008.

\bibitem[\protect\citeauthoryear{Narodytska and
  Kasiviswanathan}{2017}]{cvprw2017_blackbox}
N.~Narodytska and S.~Kasiviswanathan.
\newblock Simple black-box adversarial attacks on deep neural networks.
\newblock In {\em 2017 IEEE Conference on Computer Vision and Pattern
  Recognition Workshops (CVPRW)}, pages 1310--1318, 2017.

\bibitem[\protect\citeauthoryear{Papernot \bgroup \em et al.\egroup
  }{2016a}]{papernot2016limitations}
Nicolas Papernot, Patrick McDaniel, Somesh Jha, Matt Fredrikson, Z~Berkay
  Celik, and Ananthram Swami.
\newblock The limitations of deep learning in adversarial settings.
\newblock In {\em Security and Privacy (EuroS\&P), 2016 IEEE European Symposium
  on}, pages 372--387. IEEE, 2016.

\bibitem[\protect\citeauthoryear{Papernot \bgroup \em et al.\egroup
  }{2016b}]{papernot2016distillation}
Nicolas Papernot, Patrick McDaniel, Xi~Wu, Somesh Jha, and Ananthram Swami.
\newblock Distillation as a defense to adversarial perturbations against deep
  neural networks.
\newblock In {\em 2016 IEEE Symposium on Security and Privacy (SP)}, pages
  582--597. IEEE, 2016.

\bibitem[\protect\citeauthoryear{Papernot \bgroup \em et al.\egroup
  }{2017}]{asia_ccs2017_Papernot}
Nicolas Papernot, Patrick McDaniel, Ian Goodfellow, Somesh Jha, Z.~Berkay
  Celik, and Ananthram Swami.
\newblock Practical black-box attacks against machine learning.
\newblock In {\em Proceedings of the 2017 ACM on Asia Conference on Computer
  and Communications Security}, ASIA CCS '17, pages 506--519, 2017.

\bibitem[\protect\citeauthoryear{Pu \bgroup \em et al.\egroup }{2018}]{V2T2018}
Yunchen Pu, Martin~Renqiang Min, Zhe Gan, and Lawrence Carin.
\newblock Adaptive feature abstraction for translating video to text.
\newblock In {\em Proceedings of the Thirty-Second {AAAI} Conference on
  Artificial Intelligence, New Orleans, Louisiana, USA}, pages 7284--7291,
  2018.

\bibitem[\protect\citeauthoryear{Ren \bgroup \em et al.\egroup
  }{2015}]{nips2015_faster_rcnn}
Shaoqing Ren, Kaiming He, Ross~B. Girshick, and Jian Sun.
\newblock Faster r-cnn: Towards real-time object detection with region proposal
  networks.
\newblock In {\em NIPS}, pages 91--99, 2015.

\bibitem[\protect\citeauthoryear{Ross and
  Doshi-Velez}{2017}]{ross2017improving}
Andrew~Slavin Ross and Finale Doshi-Velez.
\newblock Improving the adversarial robustness and interpretability of deep
  neural networks by regularizing their input gradients.
\newblock {\em arXiv preprint arXiv:1711.09404}, 2017.

\bibitem[\protect\citeauthoryear{Villegas \bgroup \em et al.\egroup
  }{2017}]{ICLR2017_honglak}
Ruben Villegas, Jimei Yang, Seunghoon Hong, Xunyu Lin, and Honglak Lee.
\newblock Decomposing motion and content for natural video sequence prediction.
\newblock In {\em ICLR}, 2017.

\bibitem[\protect\citeauthoryear{Vinyals \bgroup \em et al.\egroup
  }{2015}]{cvpr2015_caption}
Oriol Vinyals, Alexander Toshev, Samy Bengio, and Dumitru Erhan.
\newblock Show and tell: A neural image caption generator.
\newblock In {\em CVPR}, 2015.

\bibitem[\protect\citeauthoryear{Wong and Kolter}{2018}]{icml2018_kolter}
Eric Wong and J.~Zico Kolter.
\newblock Provable defenses against adversarial examples via the convex outer
  adversarial polytope.
\newblock In {\em {ICML}}, 2018.

\bibitem[\protect\citeauthoryear{Xing \bgroup \em et al.\egroup
  }{2018}]{deformable2018_zhu}
Xianglei Xing, Ruiqi Gao, Song-Chun Zhu, and Ying~Nian Wu.
\newblock Deformable generator network: unsupervised disentanglement of
  appearance and geometry.
\newblock In {\em ICML workshop on Theoretical Foundations and Applications of
  Deep Generative Models (TADGM)}, July 2018.

\end{thebibliography}
}
\section{Supplementary Material: Robust feature embedding against attacks}
\begin{figure*}[ht]
	\centering
	\includegraphics[width=0.9\linewidth]{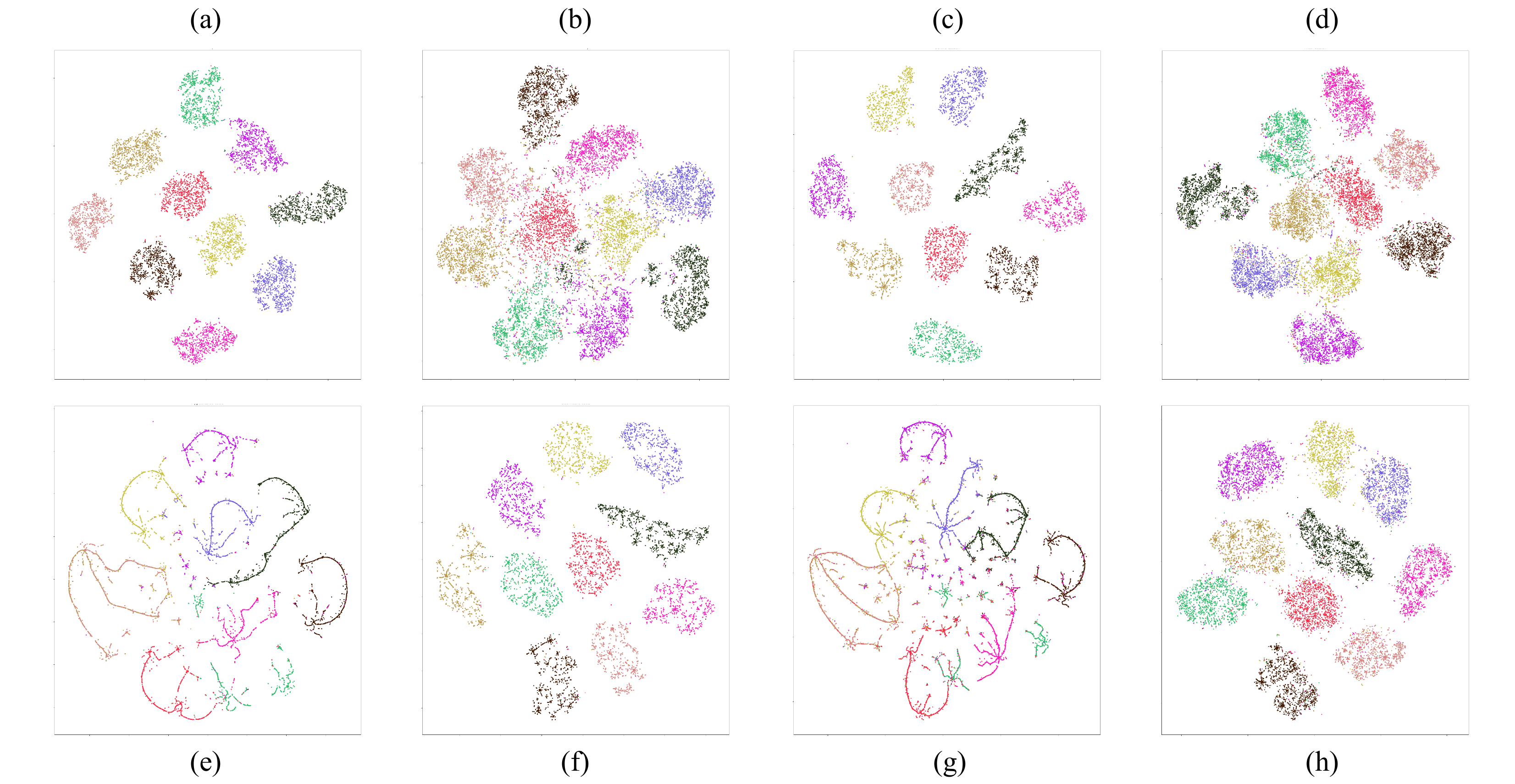}
	\caption{t-SNE visualization of the feature embedding of 
		(\textbf{a}) the baseline CNN before attack. 
		(\textbf{b}) the baseline CNN after attack. 
		(\textbf{c}) the entire code of the proposed model before attack.
		(\textbf{d}) the entire code of the proposed model after attack. The t-SNE
		(\textbf{e}) the appearance code of the proposed model before attack.
		(\textbf{f}) the geometric code of the proposed model before attack.
		(\textbf{g}) the appearance code of the proposed model after attack.
		(\textbf{h}) the geometric code of the proposed model after attack. Visualization is conducted on MNIST dataset with FGSM ($\epsilon$ = 0.3). Each color corresponds to one digit. In the visualization effect of model after attack, the square points denote the embedding code before attack and the triangle points denotes the code after attack (visible after zooming in).}
	\label{fig:tSNE}
\end{figure*}

To verify that our model indeed has a more robust feature embedding, we used t-SNE ~\cite{maaten2008visualizing} to visualize the embedding of adversarial images against legitimate images from MNIST dataset for both baseline model and our model in Figure ~\ref{fig:tSNE}. Indeed, the embedding in the joint appearance and geometric code in our model for attacked images (\ref{fig:tSNE}d) maintains stronger separability compared to baseline CNN (\ref{fig:tSNE}b). 
One of the reasons that we expect our model to be more robust against adversarial attack is that appearance and geometric features may each be more suitable for different classification tasks. Take digits as examples, several digits share the same topological structure but mainly differ in their geometric properties. By separating the feature space to $\mathbf{C}_a$ and $\mathbf{C}_g$, we provide to the classifier $\mathbf{F}_{CLS}$  shape codes $\mathbf{C}_g$ which live on a lower dimensional space than needed for traditional CNN, which makes the training more data-efficient. Interestingly, this is also reflected in the embedding structures: as shown in Figures \ref{fig:tSNE}e and \ref{fig:tSNE}g, even before attack, manifolds of different digits are entangled in the space of $\mathbf{C}_a$. But as shown in 3f and 3h, the embedding in $\mathbf{C}_g$ stays highly separable even after attack, suggesting the classifier indeed is able to utilize the more robust code $\mathbf{C}_g$.

\end{document}